\RequirePackage{fix-cm}
\documentclass{svjour3}                     
\smartqed  
\AtBeginDocument{%
  \setlength{\oddsidemargin}{\dimexpr(\paperwidth-\textwidth)/2-1in}%
  \setlength{\evensidemargin}{\oddsidemargin}%
  \setlength{\topmargin}{%
    \dimexpr(\paperheight-\textheight)/2-\headheight-\headsep-1in}%
}

\usepackage{graphicx}
\usepackage[T1]{fontenc}
\usepackage[latin9]{inputenc}
\usepackage{amsmath}
\usepackage{amssymb}
\usepackage{booktabs}
\usepackage{float}
\usepackage{enumerate}
\usepackage{enumitem}
\usepackage{url}

\usepackage{breakurl}
\usepackage[breaklinks]{hyperref}

\makeatletter
\let\cl@chapter\undefined
\makeatletter
\usepackage{cleveref}

\usepackage{subcaption}
\usepackage{graphicx}
\hypersetup{ 
    colorlinks=true, 
    linkcolor=blue,
    filecolor=blue,      
    urlcolor=blue,
    citecolor=blue
}

\usepackage{bbm}
\usepackage{multirow}
\usepackage[nolist,nohyperlinks]{acronym}
\acrodef{AI}{artificial intelligence}

\usepackage[normalem]{ulem}

\usepackage{cite}

\begin{document}

\title{The social dilemma in artificial intelligence development and why we have to solve it}
\titlerunning{The social dilemma in AI development and why we have to solve it}

\author{Inga~Str{\"u}mke    \and
       Marija~Slavkovik    \and
       Vince~I.~Madai      
       }

\institute{Inga Str{\"u}mke \at
            Department of Engineering Cybernetics, NTNU, Trondheim, Norway\\
            Department of Holistic Systems, SimulaMet, Oslo, Norway\\
            \email{inga.strumke@ntnu.no}
            \and
            Marija Slavkovik \at
            Department of Information Science and Media Studies, University of Bergen, Norway 
            \and
            Vince I. Madai \at
            QUEST Center for Transforming Biomedical Research, Berlin Institute of Health (BIH), Charit{\'e} Universit{\"a}tsmedizin Berlin, Berlin, Germany\\
            Charit{\'e} Lab for AI in Medicine, Charit{\'e} Universit{\"a}tsmedizin Berlin, Berlin, Germany\\
            School of Computing and Digital Technology, Faculty of Computing, Engineering and the Built Environment, Birmingham City University, Birmingham, United Kingdom
}

\date{Accepted for publication in Springer AI and Ethics}

\maketitle

\begin{abstract}
    While the demand for ethical artificial intelligence (AI) systems increases, the number of unethical uses of AI accelerates, even though there is no shortage of ethical guidelines. We argue that a possible underlying cause for this is that AI developers face a social dilemma in AI development ethics, preventing the widespread adaptation of ethical best practices. We define the social dilemma for AI development and describe why the current crisis in AI development ethics cannot be solved without relieving AI developers of their social dilemma. We argue that AI development must be professionalised to overcome the social dilemma, and discuss how medicine can be used as a template in this process.
\end{abstract}

\section{Introduction}\label{sec:introduction}
A professional should not have to choose between their job and doing the right thing. Still, \ac{AI} developers can be and are put in such a position. Take the example of a company that develops an \ac{AI} tool to be used to guide hiring decisions: After the product has reached a certain stage, developers may identify ethical challenges, e.g.\ recognising that the tool is discriminatory against minorities. Avoiding this discrimination may require decreasing the performance of the product. What should the developers do to rectify the situation? They necessarily need to inform management about their concern, but their complaint can be met with indifference, and even a threat to replace them\footnote{This example is a generalization of numerous experiences the authors have of being approached at relevant conferences by developers who perceive their work as unethical, e.g. as discriminatory against minorities. A common question within this context is whether they should risk losing their jobs for prioritising ethical considerations}. 

Situations such as these fall into the category of {\em social dilemmas}, and our goal in this paper is to highlight the impediment to ethical \ac{AI} development due to the social dilemma faced by \ac{AI}developers. We argue that the current approaches to ethical practices in \ac{AI} development fail to account for the existence of the challenge for developers to choose between doing the right thing and keeping their jobs.

A social dilemma exists when the best outcome for society would be achieved if everyone behaved in a certain way, but actually implementing this behaviour would lead to such drawbacks for an individual that they refrain from it. The problem we identify is that the current structures often put the burden to refuse unethical development on the shoulders of the developers who cannot possibly do it, due to their social dilemma. Furthermore, this challenge will become increasingly relevant and prevalent since \ac{AI} is becoming one of the most impactful current technologies, with a huge demand for development~\cite{economicAIimpact,ITU2018}. 

Advances in the field of \ac{AI} have led to unprecedented progress in data analysis and pattern recognition, with subsequent advances in the industry. This progress is predominantly due to machine learning, which is a data-driven method. The utilized data are in the majority of cases historical, and can thus represent discriminatory practices and inequalities. Therefore, many machine learning models currently in use cement or even augment existing discriminatory practices and inequalities. Furthermore, \ac{AI} technology does not have to be discriminatory for its development to be unethical. Mass surveillance, based on e.g.\ facial recognition, smart policing and safe city systems, are already used by several countries~\cite{ai_surveillance}, news feed models used by social media create echo chambers and foster extremism~\cite{cinelli2020echo}, and autonomous weapon systems are in production~\cite{AI_arms_race}. 

There has been rapid development in the field of \ac{AI} ethics, and sub-fields like machine learning fairness, see e.g.~\cite{cond_independence,fairness_explained,fairmlneurips,fairmlbook}. However, it is not clear that much progress is made in implementing ethical practices in \ac{AI} development, nor that developers are being empowered to refuse engaging in unethical \ac{AI} development. Reports such as The AI Index 2021 Annual Report~\cite{aiindex} stress the lack of coordination in \ac{AI} development ethics. Specifically, one of the nine highlights in this report states that ``\ac{AI} ethics lacks benchmarks and consensus''. 

Major corporations, also referred to as ``Big Tech'', are the ones developing the overwhelming majority of \ac{AI} systems in use. These corporations have reacted to academic and public pressure by publishing guidelines and principles for \ac{AI} development ethics. There has been what can be characterised as an inflation of such documents over the past years~\cite{aiethics_landscape,ethics_of_aiethics,aiethics_policy}. Although researchers and the society view \ac{AI} development ethics as important~\cite{AIethics_community}, the proliferation of ethical guidelines and principles has been met with criticism from ethics researchers and human rights practitioners who, e.g., oppose the imprecise usage of ethics-related terms~\cite{Floridi2019Unified,AIethics_toothless}. The critics also point out that the aforementioned principles are non-binding in most cases and due to their vague and abstract nature also fail to be specific regarding their implementation. Finally, they do not give developers the power to refuse unethical AI development. The late firings of accomplished \ac{AI} ethics researchers~\cite{gebru1,gebru2,gebru3} for voicing topics inconvenient for the business model of the employer, demonstrate that top-down institutional guidelines are subject to executive decisions and can be overruled. 
While we acknowledge that we must be cautious when generalizing from single cases, we are not alone with our concern that ethical principles might be merely ethics washing~\cite{ethics_washing1,ethics_washing2,ethics_washing3}, i.e., that corporations only give the impression of ethical practices in order to avoid regulation. Thus, the need for implementing ethical principles in \ac{AI} development remains, and a crucial factor for this to succeed is removing the social dilemma for \ac{AI} developers.
 
Social dilemmas exist in most areas where individuals, employers and society are in a relational triangle around decisions that affect the society at large. \ac{AI} is not an exception; there are many fields  that encounter social dilemmas and some have successfully implemented mitigating measures. A very prominent example of this is medicine. In this paper, we argue that medicine's strong focus on professionalization and the development of binding professional ethical codes is a powerful way to protect medical professionals from social dilemmas, and we discuss how structures like those in medicine can serve as a blueprint for \ac{AI} development, thus leading to a lasting impact on ethical \ac{AI} development.

Clearly, the issue of how to ensure ethical conduct from the for profit products developed by the for profit companies cannot be reduced to resolving the issue of the social dilemma  for the employees. This is a complex question that relies on constructing legal frameworks to address  ``big tech'' regulation and it is one of broad public interest~\cite{WSJ2021}. In this paper we focus only on the bottom-up contribution to the resolution of this complex problem, but recognise that the top-down legal regulation is a necessary component as well. 

Before proceeding, we recognise that our analysis touches upon topics from other ethics sub-fields, namely business ethics, corporate ethics and research ethics. We do not adopt the viewpoint of any of these since we believe that our analysis can inform them and would be hampered by a too narrow focus.

Our aim is to raise awareness towards the existence of the social dilemma for \ac{AI} developers, and by doing so outline the need for a systematic solution that removes that social dilemma. This process is a societal task, and will require interdisciplinary expertise from other fields outside of \ac{AI} development.

The paper is structured as follows. In~\cref{sec:socdil}, we carefully define the social dilemma in \ac{AI} development and highlight how it differs from known instances of social dilemmas. In~\cref{sec:codleg}, we elaborate on why professional codes of behaviour supplement legislation in tackling the serious social problem that is the regulation of technology impact. In~\cref{sec:codmed}, we take a lesson from the field of medicine on how social dilemmas faced by medical professionals are removed by establishing a professional code of conduct. In~\cref{sec:towards}, we discuss issues with establishing ethical codes of conduct for AI developers, as we see them today, and in~\cref{sec:related_work} we discuss related work. Finally, we outline our conclusions in~\cref{sec:concl}. 
\section{The social dilemma in AI development}\label{sec:socdil}
A social dilemma, also referred to as a `collective action problem', is a decision-making problem faced when the interests of the collective conflict with the interests of the individual making a decision. It was established in the early analysis of the problems of public good cost by~\cite{perrow1965,perrow1971}, who stated that ``rational self-interested individuals will not act to achieve their common or group interests.''
Well known problems that can be considered instances of social dilemmas are the prisoner's dilemma~\cite{LuceR1957}, the tragedy of the commons~\cite{Lloyd1980}, the bystander-problem~\cite{DarleyL1968}, fishing rights, et alia.
The best known of these is perhaps the tragedy of the commons, which is a situation in which individuals with open access to a shared resource selfishly deplete the resource, thus acting against the common good and hurting their own individual interests as a result.
All collective action problems concern situations in which individuals fail to behave according to the interests of the collective, although this would ultimately benefit all individuals, or, as stated by~\cite{kollock1998}: ``situations in which individual rationality leads to collective irrationality''.
At the same time, all these examples are metaphors that stand as evidence for the difficulty of formulating an exact definition of social dilemmas~\cite{AllisonBM1996}.

In the context of \ac{AI}, the social dilemma has been little discussed. The exception is in relation with autonomous vehicles~\cite{Bonnefon2016}. Bonnefon et al \cite{Bonnefon2016} observe in their experiments that ``people praise utilitarian, self-sacrificing AVs and welcome them on the road, without actually wanting to buy one for themselves.'', and  state that this has ``\dots the classic signature of a social dilemma, in which everyone has a temptation to free-ride instead of adopting the behavior that would lead to the best global outcome.''. This is, in fact, the tragedy of the commons~\cite{Hardin1986}. 

The social dilemma in \ac{AI} development described in the introduction, however, does not fit the metaphor of the tragedy of the commons, or any of the other commonly used social dilemma metaphors. Consequently, we need to define the social dilemma in the context of \ac{AI} development, and put forward the following definition:
\textit{a social dilemma exists when the best outcome for society would be achieved if everyone behaved in a certain way, but actually implementing this behaviour would lead to such drawbacks for individuals that they refrain from the behaviour}. In the social dilemma in \ac{AI} development, we encounter three agents, each with their, possibly conflicting, interests: society, a business corporation, and an \ac{AI} developer who is a member of society and an employee of the business corporation. The interest of society is ethical \ac{AI} development;
the interest of the business corporation is profit and surviving in the market; the interest of the developer is primarily maintaining their employment, but secondly ethical \ac{AI} development, because developers are also a part of society. The developer is thus put in a situation where they have to weigh their interest as a member of society and their interests as an employee of the corporation. This is the social dilemma we want the \ac{AI} developer {\bf not} to face. 

An analysis by PricewaterhouseCoopers~\cite{sizing_the_price} stated that \ac{AI} has the potential to contribute $15.7$ trillion dollars to the global economy by 2030. This puts business corporations in a competitive situation, especially  regarding developing and deploying \ac{AI} solutions fast. Fast development is potentially the opposite of what is needed for ethical development, which can require decreasing the development speed to implement necessary ethical analyses, or even deciding against deploying a system based on ethical considerations. This can create a direct conflict between the corporations' motivation and the interest of society, which manifests in the work and considerations of the developers. These then find themselves in a situation where they might be replaced if they voiced concerns or refuse to contribute to the development. 

The intention is not for AI developers to be the centre of decision making processes about what is ethical. There is a consensus \cite{CharisiDFLMSSWY17,StandardP7001,IEEE7000-2021} that the decision of what is ethical should be one taken as an agreement among all identified stakeholders in a society in which the AI system will  be developed and deployed. However, what is frequently neglected is the specification of how that agreement on what is ethical should be reached. As Baum \cite{Baum20} elaborates, collectively deciding what is ethical is not a simple process. Since no mechanisms to reach a stakeholder agreement are put in place,  developers are being put in a position to be the judge and jury on what is ethical, without having been trained at working with wider communities to achieve a collective understanding of the moral and societal impact of the system they are building. But even if they are trained for the task, their position in the company does not necessarily empower them to act upon it. 
This situation is reminiscent of the so-called principal-agent problem: The AI developers are agents of the principals in the form of their employers, and the social dilemma situation constitutes a conflict between the AI developers and their employers~\cite{jensen1976,Sabel2004,Akerlof2005}.

Expecting that \ac{AI} developers will overcome this social dilemma without support is unrealistic. This stance is strengthened by the observation of other areas where social dilemmas are evident, e.g.\ climate change, environmental destruction, and reduction of meat consumption, where billions of people behave contrary to the agreed-upon common goal of sustainability, because of their social dilemmas.  

\ac{AI} development ethics, however, are much more complex than for example the ethics of meat consumption. The ethical challenges in \ac{AI} are often both novel and complicated, with unforeseeable effects. While different approaches to ethics may not provide the same answer to the question ``What is ethical development?'', the process of analysing the ethical aspects of a development process or system yields important information regarding the risks that can be mitigated by the developer. Yet, analysing a system and its potential impact from an ethical standpoint requires ethical training and a methodology. For the \ac{AI} developer untrained in ethics and facing a social dilemma, it is unrealistic to perform this task, especially at scale~\cite{mclennan_embedded_2020}.

We can also observe the potential for a social dilemma to occur on another level, this time for corporations: No single corporation or small group of corporations can take on the responsibility of solving \ac{AI} development ethics, as this might put them at a disadvantage compared to other agents in the same market\footnote{We acknowledge that there might be cases where ethical development of a product can be considered a competitive business advantage. The existence of such cases does not preclude, however, that also cases exist where ethical development is a clear disadvantage.}. It is an interesting phenomenon that the social dilemma spirals upwards, in the sense that it can only be removed by solving it at the lowest level. If no corporation finds developers willing to engage in unethical development, they cannot end up in the corporation-level social dilemma. Furthermore, imposing corporation-level regulations for ethical conduct would likely lead to a search for loopholes, especially since there would always be gray zones, context-dependence and need for interpretation. From this perspective, solving the social dilemma for developers is also the approach that would lead to the most stable solution.
\section{Professional codes versus legislation} \label{sec:codleg}
Ethical perspectives in \ac{AI} development are important since unethical development of \ac{AI} can have a profound, negative impact both on individuals and society at large.
Motivated by recent efforts to propose a regulatory framework for \ac{AI}~\cite{AI_regulation_EU}, one might be tempted to think that the challenge of ethical \ac{AI} development could be solved solely by legislation. However, there are several reasons why legislation cannot fill this role: Legislation develops at a much slower pace than current technology, implying that legislation is likely to arrive after harm has already been done, or even worse, after customary practice has been established. Furthermore, legislating against anything that could potentially be unethical or misused would disproportionately hinder progress, which is both undesirable and would in practice affect small businesses more than large ones, reinforcing the already problematic power imbalance between users and providers. 

Note that we do not argue against legislative regulation. We are convinced that regulation is an important part of the overall approach towards ethical AI development. So are alternative approaches to AI governance such as human rights-centered design~\cite{smuha_beyond_2020,mcgregor_international_2019,yeung_ai_2020}, AI for social good~\cite{floridi_how_2020}, algorithmic impact assessment~\cite{castets-renard_human_2021}, or Ubuntu~\cite{mhlambi_rationality_2020}. We argue, however, against the notion that regulation and these alternative approaches suffice to create a stable solution. They have a blind spot and we thus face the challenging situation where - for this blind spot - we have to entrust corporations developing \ac{AI} to take the ethical responsibility, despite not being motivated purely by the benefit of society. If the corporations do not shoulder this ethical responsibility,  individual developers will be hindered in pursuing ethical development due to their social dilemma. We now describe a possible solution to this problem, recognising that the described phenomenon is not novel from a societal point of view.

Historically, societies have understood early that certain professions, while having the potential to be valuable for society, require stronger oversight than others due to their equally substantial potential for harm. While the necessity of a certain autonomy and freedom for professionals is acknowledged, it is important to simultaneously expect professionals to work for the benefit of society. As~\cite{frankel_professional_1989} puts it: ``Society's granting of power and privilege to the professions is premised on their willingness and ability to contribute to social benefit and to conduct their affairs in a manner consistent with broader social values''. Camenisch \cite{camenisch_grounding_1983} even argues that the autonomy of a profession is a privilege granted by society, which in turn leads to a moral duty of a profession to work for societal benefit. Professional codes that are not in line with societal good will be rejected by society \cite{jamal1995}.

Professional codes have been used to promote the agreed upon professional values in areas where legislative solutions are inadequate. Members of a profession are tied together in a ``major normative reference group whose norms, values, and definitions of appropriate [professional] conduct serve as guides by which the individual practitioner organizes and performs his own work''~\cite{pavalko_sociology_1988}. 
Most importantly, in the context of this work, professional codes are a natural remedy against social dilemmas encountered in professional settings. The individual is relieved from the potential consequences of criticising conduct or refusing to perform behaviour in violation of their professional codes, and it would be highly unlikely that another member of the same profession would be willing to perform the same acts in violation of the professional code. Furthermore, the public would have insight into what is the standard ethical conduct for the entire profession. 

Naturally, professional codes do not develop in a void. They draw from ethical theories, the expectations of society and from the self-image of the professionals. Consequently, professional codes are never set in stone but are constantly revised in light of technical advancement, development in societal norms and values, and regulatory restrictions. However, although they are dynamical, there is still at any given time a single version that is valid, protecting the individual professional from the social dilemma and maximizing the benefit for society.

We acknowledge that the development of professional codes is not an easy task. We thus argue that it is best to draw from a field that has succeeded at the task, as described in the next section.
\section{Professional codes in medicine}\label{sec:codmed}
As stated in the Introduction (\cref{sec:introduction}), we suggest using medicine as a template for a professional code for \ac{AI} development ethics. Although other fields have also developed professional codes, we argue based on societal impact that medicine is the most suitable example to follow. Medicine -- primarily responsible for individual and public health -- has a tremendous impact on society, at a level which few, if any, other professions share. \ac{AI} has the potential of a similarly or even more substantial impact, depending on future development. 

Medicine is an ancient profession, with the first written records dating back to Sumeria 2000 BC~\cite{biggs_medicine_1995}. The Hippocratic oath, the first recorded document of medical professional codes, was introduced by the ancient Greeks, and its impact was so large that many laypeople, incorrectly, believe that it is still taken today~\cite{zwitter_ethical_2019}. 
The British and American medical associations drafted their first codes for ethical conduct in the 19th century~\cite{backof_historical_1991}. 
Modern medicine has evolved considerably over the past 150 years, and milestones in the development of professional codes have been the declarations of the World Medical Association~\cite{gillon_medical_1985}, which states promoting ethical professional codes as one of its main goals. The two most prominent declarations are the declarations of Geneva as a response to the cruelties performed by medical professionals in Germany and Japan during World War II~\cite{declaration_geneva};  and the declaration of Helsinki for ethical conduct of medical research~\cite{declaration_helsinki}. These documents are continuously updated and received further refinement especially after disastrously unethical events, e.g.\ the revelation of the Tuskagee Syphilis study~\cite{chadwick_historical_1997}. 
Based on these documents, professional medical associations around the world have drafted professional ethical codes. Importantly, these codes are specifically designed to not be dependent on legislation which can highly differ between countries~\cite{gillon_medical_1985}.
%
%

Medical professionals are guided in their work by these ethical codes, and are protected from the social dilemma as the publicly known ethos enables them to refuse unethical behaviour without the fear of repercussions\footnote{We do of course not claim that this system is foolproof and can prevent the social dilemma fully. The professional codes in medicine are, however, arguably the ones that protect their professionals the best in an area with highest societal impact.}. Due to the similar level of expected impact on society, we view medicine as a suitable template for a professional code for \ac{AI} development ethics. In the following section we outline how the field of \ac{AI} needs to adapt in order to develop robust, impactful, and unified professional codes in analogy to medicine. 
\section{Towards professional codes in AI}\label{sec:towards}
In this section we discuss the present issues with establishing ethical codes of conduct for AI developers and outline some possible paths towards establishing them. 
\subsection{Current issues}\label{sec:issues}
The topic of professional codes for \ac{AI} has raised considerable interest during recent years, and several works have pointed out the fluid nature of this field and its complexity, see e.g.~\cite{larsson_2020,boddington_towards_2017}. Yet, we observe that there is little tangible practical impact, in the sense that there are no broadly accepted professional codes today. We believe that this can be attributed to two major reasons:

Primarily, current analyses accept many boundary conditions as given, instead of suggesting how to change these conditions. We exemplify this as follows. It is true that even if large organizations, such as the IEEE, adopted professional codes of \ac{AI} development, a plethora of challenges would remain. Who is an \ac{AI} developer? What is the incentive for someone to join these organizations and abide by the codes? What keeps the organisation from diverging from the code if the potential gain or cost avoided is substantial? Even worse, several competing organizations might publish different codes, fragmenting the field and making it impossible for the developers and the public to know the norms of the profession. Lastly, such efforts might even be hijacked by Big Tech: They could ``support'' certain organizations in publishing ethical codes, with significant influence on content, essentially leading to a new form of ethics washing. We agree that under these boundary conditions, an implementation of professional codes for \ac{AI} seems difficult or even impossible. However, we argue that we must distinguish between an analysis of a situation and practical suggestions regarding how conditions can and should be changed. In brief, we argue that analyzing a situation will not change it; only changing relevant determining factors will.

Secondly, as long as current initiatives do not free developers from the social dilemma outlined earlier, an implementation of professional codes will inevitably fail. For example, recent work has addressed how embedding ethicists in development teams  might support ethical \ac{AI} development ~\cite{mclennan_embedded_2020,mclennan_nature_2020}. However, if these ethicists are themselves just other employees of the same corporation, the social dilemma applies to them as well.
\subsection{Possible ways forward}
\begin{figure}[htb]
    \captionsetup[subfigure]{justification=centering}
    \begin{subfigure}{0.49\textwidth}{
        \centering
        \includegraphics[width=0.95\textwidth]{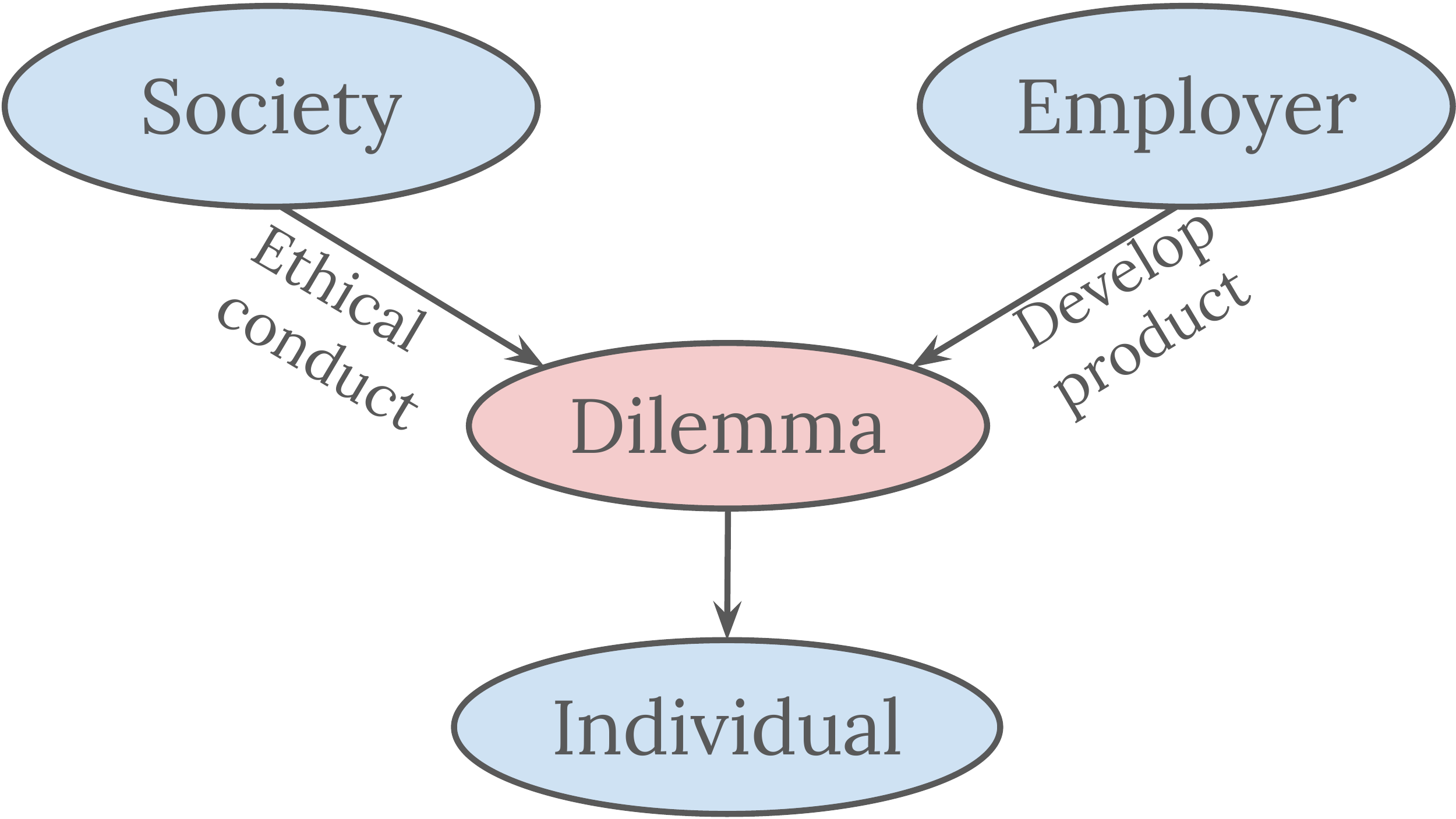}
        \caption{\label{fig:dilemma1}}}
    \end{subfigure}
    \begin{subfigure}{0.49\textwidth}{
        \centering
        \includegraphics[width=0.95\textwidth]{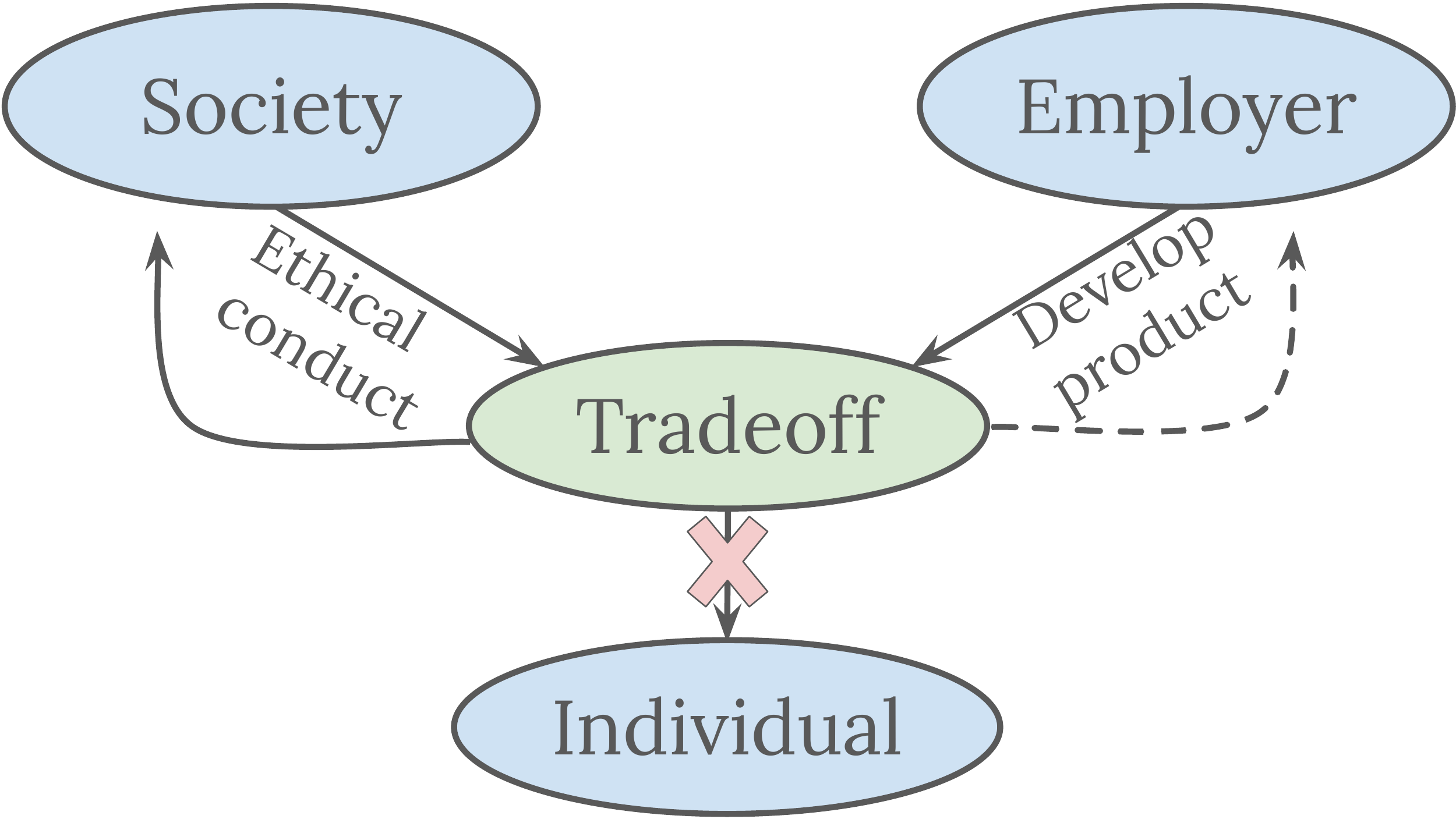}
        \caption{\label{fig:dilemma2}}}
    \end{subfigure}
    \caption{\label{fig:dilemma}
    (\protect\subref{fig:dilemma1}) Now: Society's need for ethical conduct and the employer's need to develop products together put the developer into a dilemma, and (\protect\subref{fig:dilemma2}) after introducing the ethos: what was previously a dilemma for the developer is now a trade-off that society, together with the employer, has to handle using established methods.}
\end{figure}

We argue that in order to solve the crisis in \ac{AI} development ethics, a process that addresses the two points in~\cref{sec:issues} must be initiated. Our primary proposition is that \ac{AI} development must become a unified profession, taking medicine as an example. And, as in medicine, it must become licensed. The licence must be mandatory for all developers of medium to high risk \ac{AI} systems, following, e.g., the Proposal for a Regulation laying down harmonised rules on artificial intelligence by the European Union ~\cite{AI_regulation_EU}. This would protect the individual developer in an unprecedented manner. The chances of being replaced by another professional would be very small, since employers would know that all \ac{AI} developers abide by the same code. Thus, \ac{AI} developers could refuse to perform unethical development without fear of the social dilemma consequences. The difference before and after introducing a professional ethos is depicted in~\cref{fig:dilemma}.

Secondly, national \ac{AI} developer organisations maintaining registers of employed \ac{AI} developers must be established, analogously to national medical societies. These, including all their members, would serve as nuclei for the development of professional codes, and be responsible for maintaining, updating and refining them. With such a system in place, understanding and following the codes - the professional ethos - would replace the need for individual formal training in the methodology of ethics, as is the case in medicine. Lastly, unethical behaviour could lead to the loss of ones license, which is a strong incentive not to take part in unethical development, even if required by an employer. Note how legislation does not influence the content of the professional codes but facilitates it by creating the right boundary conditions.

In his 1983 work discussing professional obligation to society~\cite{professional_ethics_abbott}, Abbott stated that one of five basic properties of professional ethics is that ``nearly all professions have some kind of formal ethical code''. While we argue that this should be the case for \ac{AI} professionals, it is not the case today. A concrete suggestion of forming an ethos along the lines of the Hippocratic oath for developers of technology was recently put forward by Abbas et al.~\cite{abbas2019hippocratic}, who suggested a Hippocratic oath for technologists, consisting of three main parts: understanding the ethical implications of technology, telling the truth and acting responsibly. The authors suggest that technologists sign the oath publicly and digitally, receiving a digital badge to be included in online profiles. The main objective of the oath being to ``raise awareness of the ethical responsibility of the technologist as a user and creator of technology'', the authors do not further define what exactly a `technologist' is, other than a creator of technology.

We do not claim that unifying \ac{AI} development into one profession is a simple task. On the contrary, we acknowledge all the challenges other authors, e.g.~\cite{mittelstadt_principles_2019}, have pointed out regarding defining who is an \ac{AI} professional, and the complex interactions between all stakeholders in \ac{AI} governance.  The difference is that we do not focus on what hinders the process, but argue 
that establishing ethical \ac{AI} development will otherwise fail: As long as professionals can be uncertain regarding whether they are an \ac{AI} developer, as long as corporations can claim that their employees are not \ac{AI} developers, as long as we leave developers alone with their social dilemma, as long as there are no single international institutions serving as contact points for governments and corporations, and as long as there is no accountability for unethical \ac{AI} development, no stable solution securing future \ac{AI} development to be ethical will be found.

Although overcoming all obstacles to a unified \ac{AI} developer profession will be a tedious endeavour, it will remove the social dilemma for developers. We argue that this is the only realistic way to ensure that \ac{AI} development follows goals in alignment with societal benefit. Once a unified profession with professional codes exists, it will serve as a safeguard against unethical corporate and governmental interests. This is important as the role of corporations can be manifold, and a unified profession will help to steer their decisions in a direction aligned with society's ethical expectations. Removing the need for internal guidelines would also remove the possibility of using \ac{AI} ethics as merely a marketing narrative. On the contrary, proven and audited adherence to professional codes could provide an economic benefit to AI companies. 

\subsection{An aside on AI as a profession}
An exact definition of who is an \ac{AI} professional is useful
to identify who  should -- or even must -- receive training in ethics for \ac{AI}. The discussion about the professionalisation of AI is outside the scope of the current discussion on the social dilemma for \ac{AI} developers. We do however, need to recognise the challenges with this point. 

Gasser et al.~\cite{gasser2020role} argue that fundamental conceptual issues such as the notion of what constitutes various ``AI professions'' remain not only open, but constitute a main challenge when discussing a single professional norm for developers of \ac{AI}. The lack of clear definitions of term `\ac{AI}' itself, what exactly a profession is and what constitutes professional ethics create, as formulated by~\cite{gasser2020role}, ``a perfect definitional storm''. A central challenge in defining who is an \ac{AI} professional, is that \ac{AI} system design is multi-disciplinary and often involve individuals that already belong to another profession, with its own professional association. \ac{AI} development can also be performed by ``those working entirely outside the framework of any professional accreditation'', as remarked in~\cite{boddington2017towards}. These aspects are precisely those we state must be overcome, acknowledging that that \ac{AI} development happens in a variety of contexts where other norms may already be relevant.

While~\cite{gasser2020role} speculate that a sufficiently strong driver for an \ac{AI} profession to evolve might emerge from a crisis -- as seen historically when modern medical ethics formed from a realisation that there was more at stake than merely individuals and their professional work~\cite{baker1999codes} -- we urge that the social dilemma for \ac{AI} developers must be resolved \textit{before} a crisis is caused by unethical \ac{AI} development.

\section{Related work}\label{sec:related_work}

The idea of a professional ethos for AI professionals and its role in achieving AI ethics has been argued for in the literature. We give a summary of some of these arguments. 
 
Stahl~\cite{Stahl2021} reviews different proposals put forward to address the challenges in the ethics of \ac{AI} on three levels: the policy-level, the organisational level and the individual level. The latter consists mainly of guidance principles and documents for individuals, designed to handle \ac{AI} systems that are already under development or in place. Stahl also points to the observation made by~\cite{Floridi2019Unified}, that the large number of guidelines for individuals can cause confusion and ambiguity, a challenge we have also addressed in this paper, and argue is best solved by developing a professional ethos for \ac{AI} professionals.

Stahl~\cite{Stahl2021}  focuses on the dilemma of control - he   stresses the importance of identifying and considering ethical issues early on during the development process, pointing out the relevance of the Collingridge dilemma~\cite{collingridge1982social}. This dilemma, also known as the dilemma of control, is the observation that ``it is relatively easy to intervene and change the characteristics of a technology early in its life cycle. However, at this point it is difficult to predict its consequences. Later, when the consequences become more visible, it is more difficult to intervene.'' This dilemma is particularly relevant for those in the position to address ethical issues during the development process. As the developers of a system are in a position to make changes to a system during the early stages, these should also be made most responsible for -- and capable of -- performing such changes. They should not be hindered by fear of repercussions, but rather encouraged by their professional responsibility. 

In another recent paper~\cite{stahl2021computer}, Stahl analyses similarities between the computer ethics discourse of the 1980s and the \ac{AI} ethics debate of today. He argues that focus should not be on the relevant ``technical artefact'', i.e.\ the computer or \ac{AI} system, but rather that ethical issues arise in the context of socio-technical systems~\cite{stahl2021computer}.He points out that ``One proposal that figured heavily in the computer ethics discourse that is less visible in the ethics of AI is that of professionalism.'', highlighting exactly the part of the discourse we argue is missing. 

Referring to literature on  ethics for computing professionals, i.a.~\cite{johnson2008computing,ethics_computing2012}, Stahl observes that  the development of professional bodies for computing has been driven exactly by the idea of                    ``institutionalising professionalism as a way to deal with ethical issues''~\cite{stahl2021computer}. 

Analysing the normative features of ethical codes in software engineering, Gogoll et al.~\cite{gogoll2021ethics} argue that codes of conduct ``are barely able to provide normative orientation in software development'', and that their value-based approach potentially prevents them from being useful from a normative perspective. Such codes being underdetermined, the authors argue that they cannot replace ethical deliberation, and rather damage the process and decrease the ethical value of the outcome. The authors instead propose to implement ethical deliberation within software development teams. This is in line with the arguments of Borenstein et al.~\cite{borenstein2021emerging}, who in their recent work on the need for \ac{AI} ethics education, discuss the fostering of a professional mindset among \ac{AI} developers. 

Discussing how \ac{AI} developers view their professional responsibilities, Gogoll et al.~\cite{gogoll2021ethics} observe that ``Oftentimes, developers believe that ethics is someone else's problem.'', conveying that \ac{AI} developers sometimes view themselves as dealing with the technology, and ethics being the responsibility of somebody else. This attitude can be seen as a symptom of social dilemmas the developers face, but do not necessarily recognise as such. Namely, it would be reasonable that the developers would be keen to avoid being put in an ethically challenging situation for which they might be held liable but are not given the tools or power to address adequately. 


The discussion on how to educate \ac{AI} professionals, including whether they should be trained in ethics, is highly relevant in the context of professionalising \ac{AI}, but somewhat outside of scope of our social dilemma for \ac{AI} developers discussion which is why we refrain from a detailed related work overview on this topic. We will mention  Borenstein et al.\cite{borenstein2021emerging} who, stressing the need for \ac{AI} ethics education, state that while many remedies to the ethical challenges resulting from \ac{AI} have been proposed, a key piece of the solution is ``enabling developers to understand that the technology they are building is intertwined with ethical dimensions, and that, as developers, they have a vital role and responsibility to engage with ethical considerations''. The authors conclude that an important part of future ethical AI is ``making sure that ethics has a central place in AI educational efforts.''. 

We do need to remark that professional education is a core offer of professional societies and thus professionalization of \ac{AI} development would in turn allow broad education about ethical issues. However, the role of continuous accreditation cannot be discounted by education alone.  A parallel can  be drawn to existing credential maintenance programs, such as for example that of the US Green Building Council's Leadership in Energy and Environmental Design (LEED) professional credentialing service\footnote{\url{https://www.usgbc.org/resources/cmp-guide)}}.

\section{Limitations}\label{sec:limit}
In the absence of the possibility to test policy interventions on societies in a randomized and controlled fashion, the decisions about the best way to achieve a certain goal, e.g.\ ethical AI development, naturally remain uncertain. We acknowledge that the impact of professional codes might be less prominent than we believe. We also acknowledge the uncertainty in whether the professionalization of AI development will lead to the desired effect that we outline. We are convinced, however, that the current debate will profit from the inclusion of the social dilemma aspect and a discussion of the potential solution that we suggest. 

\section{Conclusion}\label{sec:concl}
\ac{AI} technology has the potential for substantial advancements but also for negative impacts on society, and thus requires assurance of ethical development. However, despite massive interest and efforts, the implementation of ethical practice into \ac{AI} development remains an unsolved challenge, which in our view renders it obvious that the current approach to \ac{AI} development ethics fails to provide such assurance. Our position is that the current, guideline-based, approach to \ac{AI} development ethics fails to have an impact where it matters. We argue that the key to ethical \ac{AI} development at this stage is solving the social dilemma for \ac{AI} developers, and that this must be done by unifying \ac{AI} development into a single profession. Furthermore, we argue that, based on observations from the mature field of medicine, a unified professional ethos is necessary to ensure a stable situation of ethical conduct that is beneficial to society. While we certainly do not claim that removing the social dilemma for \ac{AI} developers is sufficient for solving all issues of \ac{AI} development ethics, we argue that it is necessary.

We have discussed ethical considerations from the perspective of added cost, but would like to also point out that ethical development has itself proper value. Awareness of ethical responsibilities both inwards (towards the corporation and peers) and outwards (towards clients and society) leads directly to the protection of assets and reputation. Professional objectives in line with ethical values leads to increased dedication and sense of ownership, resulting in higher quality deliverables. Practice in ethical consideration and evaluation processes improves professionals' decision making and implementation abilities, making them more  willing to adapt to changes required for sustainability. Focus on ethical considerations fosters a culture for openness, trust and integrity, which again decreases the risk of issues being downplayed. Outstanding professionals with the privilege to choose among several employers are likely to consider not only the opportunity for professional growth, but also whether they can expect their future employer to treat them and their peers justly and ethically.

By focusing on the social dilemma we have added additional pressure to motivate the development of professional codes for \ac{AI} ethics. Much remains to be done to operationalise this desired professional certification framework.

We can observe that the medical professional ethical code is built on a long-standing tradition of professional codes. In the field of \ac{AI}, we do not have the benefit of such a historical and globally recognised entity. Thus, the first step will be to agree on the core values and principles that apply to any \ac{AI} developer in any context. The next step will be to operationalise those values and principles on a national level by establishing a certification framework for \ac{AI} developers. Governments do not need to be left on their own when developing this certification frameworks, as it can be based on the experience with many national medical certification frameworks.

\begin{acknowledgements}
We thank Dr.\ Daniel Strech, Dr.\ Michelle Livne and Dr.\ Nora A. Tahy for the thorough review of our manuscript.
\end{acknowledgements}

\bibliographystyle{spmpsci}
\bibliography{bibliography}
\end{document}